# Causal inference and causal explanation with background knowledge

**Christopher Meek**
Department of Philosophy
Carnegie Mellon University
Pittsburgh, PA 15213*

## Abstract

This paper presents correct algorithms for answering the following two questions; (i) Does there exist a causal explanation consistent with a set of background knowledge which explains all of the observed independence facts in a sample? (ii) Given that there is such a causal explanation what are the causal relationships common to every such causal explanation?

## 1 INTRODUCTION

Directed acyclic graphs have had a long history in the modeling of statistical data. One of the earliest uses is by Sewall Wright (1921) under the name path analysis. More recently there has been a resurgence of the use of directed acyclic graphical models in statistics and artificial intelligence including work on Bayesian networks, factor analysis, and recursive linear structural equation models. The relationship between directed graphs and (sets of) distributions under a variety of assumptions has been worked out in detail by Pearl (1988), Lauritzen et al. (1990) and Spirtes et al. (1993).

A few of the benefits of using directed graphical models without latent variables include (i) the existence of direct estimates (i.e. do not need to use iterative methods for maximum likelihood estimation), (ii) the models can represent many joint distributions with a reduction in the number of parameters as compared to the number of parameters required for an unconstrained model, and (iii) the existence of efficient algorithms for calculation of conditional distributions.

An additional benefit of the directed graphical framework is that there is often a natural causal interpretation to the graphical structure. Both Pearl and Verma (1991) and Spirtes et al. (1993) have advanced theories relating causality, directed graphs and probability measures and have developed algorithms for inferring causal relationships from statistical data.

In this paper, I extend this work on causal inference to consider the following two types of questions. (i) Does there exist a causal explanation consistent with a set of background knowledge which explains all of the observed independence facts in a sample? (ii) Given that there is such a causal explanation what are the causal relationships common to every such causal explanation? A special case of the first question, where there is no background knowledge, has been answered in Verma and Pearl (1992). I consider the more realistic case where the modeler may have additional information about causal relationships. The source of the background knowledge may be prior experience of the existence or non-existence of a causal relationship, or knowledge of temporal ordering among the variables. Question (ii) is a fundamental question about the extent to which causal relationships can be inferred from a set of independence facts given the assumptions relating directed graphs, causality, and probability measures hold.

### 1.1 DEFINITIONS

A *dependency model* is a list $\mathcal{M}$ of conditional independence statements of the form $\mathbf{A}\perp\!\!\!\perp\mathbf{B}|\mathbf{S}$ where $\mathbf{A}$, $\mathbf{B}$, and $\mathbf{S}$ are disjoint subsets of $\mathbf{V}$.[1] $\mathcal{M} \models \mathbf{A}\perp\!\!\!\perp\mathbf{B}|\mathbf{S}$ if and only if $\mathbf{A}\perp\!\!\!\perp\mathbf{B}|\mathbf{S}$ appears in list $\mathcal{M}$. A graph is a pair $\langle \mathbf{V}, \mathbf{E}\rangle$ where $\mathbf{V}$ is a set of vertices and $\mathbf{E}$ is a set of edges. A *partially directed graph* is a graph which may have both undirected and directed edges and has at most one edge between any pair of vertices. A partially directed graph is said to be *directed* if and only if there are no undirected edges in the graph and a partially directed graph is *undirected* if and only if there are no directed edges in the graph. $A \to B$ if and only if there is a directed edge between $A$ and $B$ and $A - B$ if and only if there is an undirected edge between $A$ and $B$. The parents of a vertex $A$ (written $pa(A)$) is the set of vertices such that there is a directed edge from the vertex to $A$. The adjacencies of vertex $A$ (written $adj(A)$) is the set of vertices which share an edge with $A$.

---

*E-mail address: cm1x@andrew.cmu.edu

[1] The statement $\mathbf{A}\perp\!\!\!\perp\mathbf{B}|\mathbf{S}$ is read $\mathbf{A}$ is independent of $\mathbf{B}$ given $\mathbf{S}$ and is equivalent to $I(\mathbf{A}, \mathbf{S}, \mathbf{B})$.

Following the terminology of Lauritzen et al. (1990), a probability measure over a set of variables **V** satisfies the *local directed Markov property* for a directed acyclic graph G with vertices **V** if and only if for every $W$ in **V**, $W$ is independent of the set of all its non-descendants conditional on the set of its parents.[2] $Markov(G)$ is the set of probability measures that satisfy the local directed Markov condition with respect to $G$. Two graphs, $G$ and $G'$ are Markov equivalent if and only if $Markov(G) = Markov(G')$. $G$ *entails* that **A** is independent of **B** given **S** (written $G \models \mathbf{A} \perp\!\!\!\perp \mathbf{B}|\mathbf{S}$) if and only if **A** is independent of **B** given **S** in every probability measure in $Markov(G)$. It is easy to show that the set of entailed independence facts for two Markov equivalent graphs are identical. The following definition is from Verma and Pearl (1992) although the name has been changed. A directed acyclic graph $G$ is a *complete causal explanation* of $\mathcal{M}$ if and only if the set of conditional independence facts entailed by $G$ is exactly the set of facts in $\mathcal{M}$.

The *pattern* for a partially directed graph $G$ is the partially directed graph which has the identical adjacencies as $G$ and which has an oriented edge $A \to B$ if and only if there is a vertex $C \notin adj(A)$ such that $A \to B$ and $C \to B$ in $G$. Let $pattern(G)$ denote the pattern for $G$. A triple $\langle A, B, C \rangle$ is an *unshielded collider* in $G$ if and only if $A \to B$, $C \to B$ and $A$ is not adjacent to $B$. It is easy to show that two directed acyclic graphs have the same pattern if and only if they have the same adjacencies and same unshielded colliders.

**Theorem 1 (Verma and Pearl 1990)** *Two directed acyclic graphs $G$ and $G'$ are Markov equivalent if and only if $pattern(G) = pattern(G')$.*

A partially directed graph $G$ *extends* partially directed graph $H$ if and only if (i) $G$ and $H$ have the same adjacencies and (ii) if $A \to B$ is in $H$ then $A \to B$ is in $G$. A graph $G$ is a *consistent DAG extension* of graph $H$ if and only if $G$ extends $H$, $G$ is a directed acyclic graph, and $pattern(G) = pattern(H)$. Let $\mathcal{K}$ be a pair $\langle \mathbf{F}, \mathbf{R} \rangle$ where **F** is the set of directed edges which are forbidden, **R** is the set of directed edges which are required; these sets will represent our background knowledge. It is possible to extend the set of background knowledge to include a partial order over the variables but this extension is not handled in this paper. Background knowledge $\mathcal{K}$ is *consistent* with graph $G$ if and only if there exists a graph $G'$ which is a consistent DAG extension of $G$ such that (i) all of the edges in **R** are oriented correctly in $G'$ and (ii) no edge $A \to B$ in **F** is oriented as such in $G$.

---

[2]See Lauritzen et al. (1990) for a comparison of a variety of alternative Markov conditions. $A$ is an *ancestor* of $B$ and $B$ is a *descendant* of $A$ if and only if $A = B$ or there is a directed path from $A$ to $B$.

## 1.2 PROBLEMS

In this paper I will consider the following four question and give algorithms for answering them;

(A) Does there exists a complete causal explanation for a set of conditional independence statements $\mathcal{M}$?

(B) Does there exists a complete causal explanation for a list of conditional independence statements $\mathcal{M}$ consistent with background knowledge $\mathcal{K}$?

(C) Given that there is a complete causal explanation for $\mathcal{M}$ what are the causal relationships common to every complete causal explanation?

(D) Given that there is a complete causal explanation for $\mathcal{M}$ what are the causal relationships common to every complete causal explanation consistent with respect to background knowledge $\mathcal{K}$?

Problems (A) and (C) are just special cases of problems (B) and (D) respectively. Verma and Pearl (1992) have given an algorithm to answer problem (A).

## 1.3 OVERVIEW OF SOLUTIONS

In this section I will outline solutions of problems (B) and (D). The algorithm for solving problem (B) consists of the following four phases.

I Examine independence statements in $\mathcal{M}$ and try to construct the pattern of some directed acyclic graph $G$. Let $\Pi_I$ be the result of Phase I.

II Try to extend $\Pi_I$ with the background knowledge $\mathcal{K}$. Let $\Pi_{II}$ be the result of Phase II.

III Try to find a graph $\Pi_{III}$ which is a consistent DAG extension of $\Pi_{II}$.

IV Check whether $\Pi_{III}$ is a complete causal explanation for $\mathcal{M}$.

The solution to problem (D) and thus problem (C) is closely related to the solution of problem (B); The algorithm to solve problem (D) consists of phase I and phase II described above. The work comes in showing that the orientation rules used in Phase II yield a graph which has the required property of having *all and only* the orientations common to complete causal explanations for $\mathcal{M}$ consistent with a set of background knowledge $\mathcal{K}$.

## 2 CAUSAL RELATIONSHIPS COMMON TO ALL COMPLETE CAUSAL EXPLANATIONS

### 2.1 Problem (C)

The solution to problem (C) consists of phase I and phase II' described below.

### 2.1.1 Phase I

The goal of phase I is to find the pattern which represents the class of complete causal explanations for $\mathcal{M}$. This is accomplished in two steps described below. A triple $\langle A, B, C \rangle$ is said to be *unshielded* if and only if $A$ is adjacent to $B$, $B$ is adjacent to $C$ and $A$ is not adjacent to $C$.

- S1 Form an undirected graph $G$ by the following rule. $A$ is adjacent to $B$ in $G$ if and only if there does not exist a set $\mathbf{S} \subseteq \mathbf{V} \setminus \{A, B\}$ such that $\mathcal{M} \models A \perp\!\!\!\perp B | \mathbf{S}$. If there is such an $\mathbf{S}$ let $Sep(A, B) = \mathbf{S}$.
- S2 For all unshielded triples $\langle A, B, C \rangle$ orient $A \to B$ and $C \to B$ if $B \notin Sep(A, C)$.

### 2.1.2 Phase II'

The goal of phase II' is to find a partially directed graph whose adjacencies are the same as any complete causal explanation for $\mathcal{M}$ and whose edges are directed if and only if every complete causal explanation for $\mathcal{M}$ has the edges oriented.

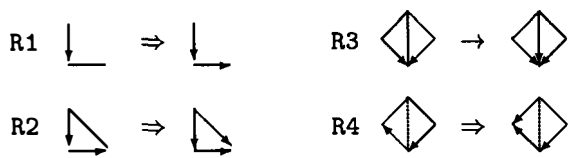

Figure 1: Orientation rules for patterns

A brief explanation of the schematic rules in Figure 1. Each orientation rule consists of a pair of schematic graphs. A schematic graph matches a pattern $\Pi'$ if there exists a set of vertices D in $\Pi'$ and a bijective mapping $(f)$ from the vertices in the schematic pattern to D such that (i) pairs of vertices are adjacent in the schematic if and only if the corresponding pair of vertices are adjacent in $\Pi'$ and (ii) if $A \to B$ in the schematic then the corresponding edge is oriented $f(A) \to f(B)$ in $\Pi'$ (iii) if $A - B$ in the schematic then the corresponding edge is unoriented and (iv) if $A$ and $B$ are connected by a dashed line then either $f(A) - f(B)$, $f(A) \to f(B)$, or $f(B) \to f(A)$ appears in $\Pi'$. If the schematic to the left of the $\Rightarrow$ matches pattern $\Pi'$ then orient the unoriented edges in $\Pi'$ according to the oriented edges in the schematic to the right of the $\Rightarrow$.

Given the rule in Figure 1 phase II' is a one step algorithm.[3]

- S1 Let $\Pi_I$ be the result of phase I. Orient every edge which can be oriented by successive applications of rules R1, R2 and R3; i.e. close $\Pi_I$ under rules R1, R2 and R3.

## 2.2 Problem (D)

The solution for problem (D) consists of phases I, II' and II''. Phase II'' is described below.

### 2.2.1 Phase II''

Let $\mathcal{K} = \langle \mathbf{F}, \mathbf{R} \rangle$ be the background knowledge and let $\Pi_{II'}$ be the partially directed graph obtained from phase II'.[4]

- S1 If there is an edge $A \to B$ in $\mathbf{F}$ such that $A \to B$ is in $\Pi_{II'}$ then FAIL.
- S1' If there is an edge $A \to B$ in $\mathbf{R}$ such that $B \to A$ is in $\Pi_{II'}$ or $A$ is not adjacent to $B$ then FAIL.
- S2 Randomly choose one edge $A \to B$ from $\mathbf{R}$ and let $\mathbf{R} = \mathbf{R} \setminus \{A \to B\}$.
- S3 Orient $A \to B$ in $\Pi_{II'}$ and close orientations under R1, R2, R3, and R4.
- S4 If $\mathbf{R}$ is not empty then go to S1.

If Phase II'' fails then there is no complete causal explanation for $\mathcal{M}$ consistent with $\mathcal{K}$.

## 2.3 Correctness

By assumption there is a directed acyclic graph $G$ which is a complete causal explanation for $\mathcal{M}$.[5] Since a graph $G'$ which is Markov equivalent to $G$ has the same entailed independence facts $G'$ is also a complete causal explanation for $\mathcal{M}$. Any graph $G'$ which is not Markov equivalent to $G$ is not a complete causal explanation for $\mathcal{M}$; either $G'$ differs from $G$ by (i) an adjacency between $A$ and $B$ in which case for some set $\mathbf{S}$ it is the case that $A \perp\!\!\!\perp B | \mathbf{S}$ is entailed in one but not the other graph and (ii) there is an unshielded triple $\langle A, B, C \rangle$ which is a oriented $A \to B$ and $C \to B$ in one but not the other graph in which case there is a set $\mathbf{S}$ which does not include $B$ such that $A \perp\!\!\!\perp C | \mathbf{S}$ in one but not the other graph. The correctness of phase I follows from the correctness of the PC algorithm (Spirtes et al. 1993) or the correctness of the algorithm presented in Verma and Pearl (1992). However, the PC algorithm is more judicious than the algorithm presented in phase I with respect to the number and type of independence facts which need to be checked which, in practice, leads to an efficient implementation with nice statistical properties (see Spirtes et al. 1993).

---

[3]This phase can be implemented in a procedure with a running time polynomial in the number of vertices in the graph. The output of this phase is a maximally oriented graph (defined below). Chickering (1995) and Andersson et al. (1995) give algorithms for finding the maximally oriented graph from a directed graph rather than a pattern. Chickering (1995) also gives an algorithm to find the maximally oriented graph from a pattern; this algorithm is more complicated but more efficient than a naive implementation of the method described above.

[4]This phase can be implemented in a procedure with a running time polynomial in the number of vertices in the graph.

[5]The existence of a complete causal explanation for $\mathcal{M}$ is equivalent to the assumption of faithfulness (Spirtes et al. 1993) or stability (Verma and Pearl 1990).

Given that the correct pattern has been found in phase I problems (C) and (D) can be restated. To solve problem (C) all of the orientations common to Markov equivalent graphs with the pattern obtained from phase I. To solve problem (D) all of the orientations common to Markov equivalent graphs with the pattern obtained from phase I with the additional restriction that the orientations agree with the edges in $\mathcal{K}$, the background knowledge. The following definition formalizes these notions.

The *maximally oriented graph* for pattern $G$ with respect to a consistent set of background knowledge $\mathcal{K} = \langle \mathbf{F}, \mathbf{R} \rangle$ is the graph $max(G, \mathcal{K})$ such that for each unoriented edge $A - B$ in $max(G, \mathcal{K})$ there exist graphs $G_1$ and $G_2$ that are consistent DAG extensions of $max(G, \mathcal{K})$ such that (i) $A \to B$ in $G_1$ and $B \to A$ in $G_2$, (ii) every edge in $\mathbf{R}$ is oriented correctly in $max(G, \mathcal{K})$, and (iii) no edge $A \to B$ in $\mathbf{F}$ is oriented as $A \to B$ in $max(G, \mathcal{K})$.

An orientation rule is *sound* if and only if any orientation other than the orientation indicated by the rule would lead to a new unshielded collider or a directed cycle.

**Theorem 2 (Orientation Soundness)** *The four orientation rules given in Figure 1 are sound.*

**Theorem 3 (Orientation completeness)** *The result of applying rules R1, R2 and R3 to a pattern of some directed acyclic graph is a maximally oriented graph.*

**Theorem 4 (Comp. w/ Back. Knowledge)**
*Let $\mathcal{K}$ be a set of background knowledge consistent with pattern $\Pi$ of some directed acyclic graph. The result of applying rules R1, R2, R3 and R4 (and orienting edges according to $\mathcal{K}$) to a pattern $\Pi$ is a maximally oriented graph with respect to $\mathcal{K}$.*

The proofs of these theorems are given in the appendix.

# 3   EXISTENCE OF COMPLETE CAUSAL EXPLANATIONS

In this section I will present solutions for problems (A) and (B). As mentioned above, Verma and Pearl (1992) gave a solution to problem (A). Their solution of problem (A) consists essentially of phase I described above and of phase III and phase IV presented below. However, phase III has been modified and their solution does not handle background knowledge (i.e. does not solve problem (B)). The modification to phase III will be described below.

## 3.1   Problem (A) and (B)

The solution to problem (B) subsumes the solution to problem (A); problem (A) is the instance of problem (B) with no background knowledge. The solution of problem (B) consists of four phases. The first two phases (phase I and phase II) have been described above and the final two (phase III and phase IV) are described below.

### 3.1.1   Phase III

Let $\Pi_{II}$ be the result of phase II. Phase III attempts to find a consistent DAG extension of $\Pi_{II}$.

S1  If $\Pi_{II}$ has no unoriented edges then STOP

S2  Choose an unoriented edge $A - B$ from $\Pi_{II}$

S3  Orient edge $A \to B$ in $\Pi_{II}$ and close orientations under rules R1, R2, R3, and R4.

S4  Go to S1.

The significant difference between this algorithm and the algorithm presented in Verma and Pearl (1992) is that their algorithm has the "potential" for backtracking. Each time that an edge is oriented in step III the edge had to be pushed onto a stack in case that the specific choice of orientation could not be extended to a consistent DAG extension of $\Pi_{II}$. They conjectured that there is no need for the backtracking on the basis of empirical studies. Their conjecture is correct; the conjecture follows from Theorem 4

### 3.1.2   Phase IV

Let $\Pi_{III}$ be the result of Phase III.

S1  If $\Pi_{III}$ is cyclic then FAIL

S2  Test that every statement I in $\mathcal{M}$ is entailed by $\Pi_{III}$ (i.e. $\Pi_{III} \models I$).

S3  Let $\prec$ be a total ordering of the nodes of $\Pi_{III}$ which agrees with the orientations in $\Pi_{III}$, i.e. $A \to B$ implies that $A \prec B$. Let $A_\prec$ be the set of vertices which are before $A$ in ordering $\prec$. For all vertices $A$ test if $\mathcal{M} \models A \perp\!\!\!\perp (A_\prec \backslash pa(A)) | pa(A)$

To test whether an independence fact is entailed by a given directed acyclic graph one can use either d-separation (Pearl 1988) or Lauritzen's rule (Lauritzen et al. 1990). The correctness of phase IV has been shown (see Verma and Pearl 1992) for dependency models $\mathcal{M}$ closed under the graphoid axioms (see Pearl 1988). An alternative approach without this restriction is to replace steps 2 and 3 with the single step of checking to see that for all disjoint subsets $\mathbf{A}$, $\mathbf{B}$, and $\mathbf{C}$ of the set of vertices it is the case that $\mathcal{M} \models \mathbf{A} \perp\!\!\!\perp \mathbf{B} | \mathbf{C}$ if and only if $\Pi_{III} \models \mathbf{A} \perp\!\!\!\perp \mathbf{B} | \mathbf{C}$.

## 3.2 Correctness

The correctness of the solution to problem (B) is shown as follows. Assume that there is a complete causal explanation for $\mathcal{M}$ with respect to $\mathcal{K}$ and that graph $G$ is such an explanation. As discussed above, phase I finds the graph $pattern(G)$ and phase II finds $max(pattern(G), \mathcal{K})$. To see that phase III finds a consistent DAG extension of $pattern(G)$ observe that after each iteration of the steps in phase III the result is a graph which is a maximally oriented graph for $pattern(G)$ and some set of background knowledge, i.e. $max(pattern(G), \mathcal{K}')$ for some $\mathcal{K}'$. Thus any choice of orientation in step 3 will be consistent and graph resulting from phase III will be a complete causal explanation for $\mathcal{M}$ with respect to $\mathcal{K}$. If contrary to our original assumption; there is no complete causal explanation for $\mathcal{M}$ with respect to $\mathcal{K}$ then in the event that the algorithm reaches phase IV with some graph $G$ then the graph will fail phase IV.

## 4 RELATED TOPICS

In this section the relationship of this work to other questions of interest in graphical modeling is briefly considered.

### 4.1 Counting principles

The proof of Theorem 3 gives us a method to calculate the number of Markov equivalent graphs in a given Markov equivalence class and the proof of Theorem 4 gives us a method to calculate the number of Markov equivalent graphs that share a set of orientation in a given Markov equivalence class. While the methods are too involved to present in the space which is available the interested reader can reconstruct these algorithm from an analysis of the proofs of these theorems.

### 4.2 Find a DAG from a pattern

Combining phases II' and III gives us an algorithm which converts a pattern $G$ to a directed acyclic graph $H$ such that $pattern(H) = G$. This is of use for at least the following reason. Information scores (MDL, AIC, and BIC) and the scores proposed by Heckerman et al. (1994) are identical for Markov equivalent models. These scores are used as the basis of model selection techniques. Phases II' and III allow a model selection procedure to use the more appropriate space of patterns to search for models during the process of model selection. One such procedure is presented in Spirtes and Meek (1995).

### 4.3 Chain graphs

A *cycle* in a partially directed graph $G$ is a sequence of vertices $\langle A_1, A_2, \ldots, A_n \rangle$ with $n > 2$ such that (i) $A_1 = A_n$, (ii) all other pairs of vertices are distinct, (iii) for all $1 \leq i < n$ it is the case that either $A_i \rightarrow A_{i+1}$ or $A_i - A_{i+1}$ is in $G$, and (iv) for some $1 \leq i < n$ it is the case that $A_i \rightarrow A_{i+1}$.

A *chain graph* is an acyclic partially directed graph. The chain graph representation subsumes both directed and undirected graphical models. A discussion of directed and undirected models can be found in Pearl (1988) and Whittaker (1990) and a discussion of chain graphs can be found in Lauritzen and Wermuth (1989), Whittaker (1990) and Frydenberg (1990).

**Lemma 1** *Let $\Pi_0$ be the result of applying the orientation rules R1, R2, and R3 to the pattern $\Pi$ of some directed acyclic graph. In $\Pi_0$, if $A \rightarrow B$ and $B - C$ then $A \rightarrow C$.*

From Lemma 1 it follows that if $\Pi$ is the pattern for a Markov equivalence class for some directed acyclic graph then the graph $max(\Pi, \emptyset)$ obtained by applying phase II' to $\Pi$ is a chain graph.[6] Thus $max(\Pi, \emptyset)$ constitutes a natural way to represent the entire Markov equivalence class of graphs with a single chain graph. This may allow researchers to develop search techniques for directed acyclic representations using chain graphs.

### Acknowledgements

I would like to thank Thomas Richardson, Clark Glymour, Peter Spirtes and two anonymous referees for helpful comments on an earlier draft of this paper. Research for this paper was supported by the Office of Navel Research grant ONR #N00014-93-1-0568.

---

[6]Andersson et al. (1995) have also shown this result in an unpublished manuscript. In Verma and Pearl (1990) $max(\Pi, \emptyset)$ is called the completed pattern.

## 5   APPENDIX — PROOFS

**Theorem 2** *The four orientation rules given in Figure 1 are sound.*

**Proof** — Rule R1; If the edge were oriented in the opposite direction there would be a new unshielded collider. Rule R2; If the edge were oriented in the opposite direction there would be a cycle. Rule R3; if the edge were oriented in the opposite direction then by two application of the rule R2 there would be a new unshielded collider. Rule R4; If the edge were oriented in the opposite direction then by two applications of rule R2 there would be a new unshielded collider. □

**Lemma 1** *Let $\Pi_0$ be the result of applying the orientation rules R1, R2, and R3 to the pattern $\Pi$ for some directed acyclic graph. In $\Pi_0$, if $A \to B$ and $B - C$ then $A \to C$.*

**Proof** — A vertex $X$ is an ancestor of vertex $Y$ with respect to $\Pi_0$ if there is a path such that every edge is directed from $X$ to $Y$ in $\Pi_0$. The orientations in $\Pi_0$ induce a partial ordering on the vertices by the following rule; $X < Y$ if $X$ is an ancestor of $Y$. With respect to this partial ordering, choose a vertex $B$ to be a minimal vertex such that there are edges $A \to B$, $B - C$ in $\Pi_0$ and $A \to C$ is not in $\Pi_0$. Note that $A \in adj(C)$ otherwise $B - C$ would be oriented by rule R1. Furthermore, $A - C$ must be unoriented; if $A - C$ is oriented $A \to C$ then we are done and if the edge is oriented $C \to A$ then $B - C$ oriented by rule R2.

*Case 1* — Edge $A \to B$ is oriented in $\Pi_0$ by rule R1. Thus there is an edge $D \to A$ such that $D \notin adj(B)$. By construction $B$ is a minimal vertex such that there are edges $A \to B$ and $B - C$ in $\Pi_0$ but $A$ meets this requirement and $A < B$. Contradiction.

*Case 2* — Edge $A \to B$ is oriented because it is part of an unshielded collider. In this case, there is an edge $D \to B$ such that $D \notin adj(A)$. If $D \notin adj(C)$ then $B - C$ would be is oriented by rule R1. If $D \in adj(C)$ and $D - C$ is unoriented then $B - C$ is oriented by R3. Suppose that $D - C$ is oriented. If $C \to D$ then $B - C$ is oriented by rule R2 else if $D \to C$ then $A - C$ is oriented $C \to A$ by rule R1 and $B - C$ is oriented by rule R2. Contradiction.

*Case 3* — Edge $A \to B$ is oriented by R3. Observe that there is an unshielded collider colliding at $B$. This case is sufficiently similar to case 2 that the proof is omitted.

*Case 4* — Edge $A \to B$ is oriented by R2. In this case there exists a vertex $D$ such that $A \to D$ and $D \to B$ are in $\Pi_0$. $D \in adj(C)$ otherwise $B - C$ is oriented by R1. Edge $D - C$ is oriented by construction ($D < B$). If $C \to D$ the $B - C$ is oriented by R2; otherwise if $D \to C$ then $A \to C$ by R2. □

An undirected graph $H$ is *chordal* if and only if every undirected cycle of length four or more has an edge between two nonconsecutive vertices on the cycle (i.e. has a chord). A total order ($<$) induces an orientation in an undirected graph $H$ by the rule that if $A - B$ is in $H$ then orient the edge $A \to B$ if and only if $A < B$. If $H$ is an undirected graph and $\alpha$ is a total ordering of the vertices in $H$ then $H_\alpha$ is the induced directed graph obtained by the rule given above. Clearly $H_\alpha$ is acyclic. A total order $\alpha$ is a *consistent ordering* with respect to $H$ if and only if $H_\alpha$ has no unshielded colliders.

**Lemma 2** *Only chordal graphs have consistent orderings.*

**Proof** — Suppose that $\alpha$ is a consistent ordering with respect to non-chordal graph $H$. Let $\langle A_1, A_2, A_3, \ldots, A_n \rangle$ be a non-chordal cycle with $n \geq 4$ in undirected graph $H$. Let $A_i$ be the largest vertex (with respect to the ordering $\alpha$) in the cycle. If $i = n$ then $A_{i+1} = A_1$ and if $i = 1$ then $A_{i-1} = A_n$. In $H_\alpha$, $A_{i-1} \to A_i$ and $A_{i+1} \to A_i$ and since the cycle is non-chordal $A_{i+1} \notin adj(A_{i-1})$. Contradiction. □

A *clique* in graph $H$ is a set of vertices such that there is an edge in $H$ between each pair of vertices in the set. A *maximal clique* is a set of vertices that is a clique and such that no superset of the set is a clique. Let $\mathcal{C}_H = \{\mathcal{C}_1, \ldots, \mathcal{C}_n\}$ denote the set of maximal cliques of graph $H$. Note that maximal cliques in $\mathcal{C}_H$ can overlap and that the union of all of the maximal cliques is the set of vertices in $H$. A *join tree* for $H$ is a tree whose vertices are in $\mathcal{C}_H$ and such that (i) Each edge $\mathcal{C}_i - \mathcal{C}_j$ is labeled by the set $\mathcal{C}_i \cap \mathcal{C}_j$, and (ii) for every pair $\mathcal{C}_i$ and $\mathcal{C}_j$ ($i \neq j$) and for every $A \in \mathcal{C}_i \cap \mathcal{C}_j$ each edge along the unique path between $\mathcal{C}_i$ and $\mathcal{C}_j$ includes label $A$. Now I state a useful result from Beeri et al. 1983.

**Lemma 3 (Beeri et al.)** *Graph $H$ is chordal if and only if $H$ has a join tree.*

A partial order $\pi$ is a *tree order* for tree T if and only if for all A and B which are adjacent in T either $\pi(A,B)$ or $\pi(B,A)$. Conceptually a tree order is obtained by choosing one node as the root of the tree and ordering vertices based on their distance from the root; all tree orderings for a tree T can be obtained in this fashion by selecting each vertex as the root.

Let $\pi_T$ be a tree ordering of the join tree T for graph H. $\pi_T$ induces a partial ordering $\prec_{\pi_T}$ on the vertices of H by the following rules; (i) if $\pi_T(C_i, C_j)$ and $C_i$ is not the minimum element of $\pi_T$ then for all $A \in C_i \backslash C_j$ and $C \in C_j \backslash C_i$ and $B \in C_j \cap C_i$ order $A \prec_{\pi_T} B$ and $B \prec_{\pi_T} C$, (ii) if $\pi_T(C_i, C_j)$ and $C_i$ is the minimum element of $\pi_T$ then for all $C \in C_j \backslash C_i$ and $B \in C_j \cap C_i$ order $B \prec_{\pi_T} C$, (iii) if $A \prec_{\pi_T} D$ and $D \prec_{\pi_T} B$ then $A \prec_{\pi_T} B$ (i.e. transitive closure of $\prec_{\pi_T}$).

Let $\pi$ be a tree ordering for join tree T of H. Note that the partial order $\prec_\pi$ on the vertices in H induced by the partial order $\pi$ only orients edges which are involved in an unshielded triple; i.e. $A \prec_\pi B$ only if there is a C such that $\langle A, B, C \rangle$ or $\langle C, A, B \rangle$ is an unshielded triple. In fact all edges involved in unshielded colliders except those edges $A - B$ where both A and B are in the minimum vertex (the "root clique") of the join tree.

A partial order $\pi_1$ is an *extension* of a partial order $\pi_2$ if and only if for all A and B such that $\pi_2(A,B)$ it is the case that $\pi_1(A,B)$.

**Lemma 4** *Let $\pi$ be a tree ordering of a join tree T for H. Any extension of $\prec_\pi$ to a total ordering is a consistent ordering for H.*

**Proof** — Let $\alpha$ be a total ordering which extends $\prec_\pi$. No unshielded collider can occur inside a clique since all triples are shielded. Let $\langle A, B, C \rangle$ be an unshielded triple (i.e. A is adjacent to B, B is adjacent to C, and A is not adjacent to C). There exists an i and j such that $A \in C_i \wedge A \notin C_j \wedge C \notin C_i \wedge C \in C_j \wedge B \in C_i \cap C_j$; if not A and C would be adjacent. By the join tree property we know that there is a unique path p between $C_j$ and $C_i$ in T.

*Case 1* — $\neg(\pi(C_i, C_j) \vee \pi(C_j, C_i))$. There must be a k such that $C_k$ is on p such that $\pi(C_k, C_i) \wedge \pi(C_k, C_j)$. We know that $A \notin C_k \vee C \notin C_k$ otherwise $\langle A, B, C \rangle$ is not unshielded since $C_k$ is a clique. Without loss of generality suppose that $C \notin C_k$. We know that $B \in C_k$ by the join tree property and since $\pi(C_k, C_j)$ it is the case that $B \prec_\pi C$ and thus $\langle A, B, C \rangle$ is not an unshielded collider in $H_\alpha$.

*Case 2* — $\pi(C_i, C_j)$ (other case is symmetric). In this case the $\langle A, B, C \rangle$ unshielded triple is oriented as a non-collider by any extension of $\prec_\pi$ to a total order since $B \prec_\pi C$. □

**Lemma 5 (Orienting chordal graphs)** *Let H be an undirected chordal graph. For all pairs of adjacent vertices A and B in H there exist total orderings $\alpha$ and $\gamma$ which are consistent with respect to H and such that $A \to B$ is in $H_\alpha$ and $B \to A$ is in $H_\gamma$.*

**Proof** — For the case where H is disconnected apply the argument to each of the disconnected components.

*Case 1* — For all i either $A \in C_i \wedge B \in C_i$ or $A \notin C_i \wedge B \notin C_i$. Let $\pi$ be a tree ordering of a join tree for H. A and B are not comparable with respect to $\prec_\pi$. Thus by Lemma 4 we simply choose two extensions of $\prec_\pi$; one with $A \prec_\pi B$ and another with $B \prec_\pi A$.

*Case 2* — There exists an i such that $A \notin C_i \vee B \notin C_i$ and $A \in C_i \vee B \in C_i$. Without loss of generality assume that $A \notin C_i \wedge B \in C_i$. Given that there is an edge between A and B there is a j such that $j \neq i$ and $A \in C_j \wedge B \in C_j$. Let $\pi_1$ be a tree ordering of a join tree for H with $C_i$ is the root and let $\pi_2$ be a tree ordering of a join tree for H with $C_j$ as the root. Then consider any extension of $\prec_{\pi_1}$ and $\prec_{\pi_2}$ to total orderings and apply Lemma 4. We are done since $B \prec_{\pi_1} A$ and $A \prec_{\pi_2} B$. □

**Theorem 3** *The result of applying rules R1, R2 and R3 to a pattern of some directed acyclic graph is a maximally oriented graph.*

**Proof** — Let $\Pi_0$ be the result of applying the orientation rules R1, R2, and R3 to the pattern $\Pi$. Given Lemma 1 no orientation of edges not oriented in $\Pi_0$ will create a cycle which includes an edge or edges oriented in $\Pi_0$ and no orientation of an edge not oriented in $\Pi_0$ can create an unshielded collider with an edge oriented in $\Pi_0$. Consider the undirected graph H, the subgraph of $\Pi_0$, obtained by removing all of the oriented edges in $\Pi_0$. H is the union of disjoint chordal graphs; suppose this is not the case. Then, by Lemma 2 all total orderings of the vertices leads to a new unshielded collider, say $\langle A, B, C \rangle$, in H. By Lemma 1, the triple $\langle A, B, C \rangle$ also forms an unshielded triple in $\Pi_0$, that is $A \notin adj(C)$ in $\Pi_0$. This is a contradiction; by assumption the graph $\Pi$ and thus $\Pi_0$ have all unshielded colliders oriented and that there is an acyclic orientation of the graph $\Pi$ with no new unshielded colliders. Finally, by applying Lemma 5 we have completed the theorem. □

Let H be a partially oriented chordal graph and let T be a join tree for H. Let $\Lambda_{ij} = C_i \cap C_j$. We define a relation $\gamma_T$ on the nodes of T, the maximal cliques of H, from the orientations in H as follows; $\gamma_T(C_i, C_j)$ if and only if (i) $\Lambda_{ij} \neq \emptyset$, (ii) for all $A \in \Lambda_{ij}$ and $B \in C_j \backslash \Lambda_{ij}$ it is the case that $A \to B$ is in H and (iii) it is not the case that for all $A \in \Lambda_{ij}$ and $B \in C_i \backslash \Lambda_{ij}$ $A \to B$ is in H. We define the partial order $\epsilon_T$ on the nodes of T as follows; (i) $\epsilon_T(C_i, C_j)$ if $\gamma_T(C_i, C_j)$ and (ii) $\epsilon_T(C_i, C_k)$ if $\epsilon_T(C_i, C_j) \wedge \epsilon_T(C_j, C_k)$. That $\epsilon_T$ is a partial order follows from the fact that T is a tree and condition (iii) of the definition of $\gamma$.

**Lemma 6** *Let T be a join tree for a partially oriented chordal graph H without any unshielded colliders and with orientations closed under rules R1, R2, R3, and*

**R4.** If there exists an unshielded triple $\langle A, B, C \rangle$ such that $A \to B$ in $H$ then for all $i$ and $j$ such that $A \in C_i \wedge B \in C_i \wedge C \notin C_i$ and $A \notin C_j \wedge B \in C_j \wedge C \in C_j$ it is the case that $\gamma_T(C_i, C_j)$.

**Proof** — The proof is in two parts; Figure 2 helps to clarify the proof. Part (i) — Show that for all $C \in C_j \setminus \Lambda_{ij}$ it is the case that $B \to C$ is in $H$. Simply apply R1 to each of the required edges. Part (ii) — Show that for all $D \in \Lambda_{ij}$ and for all $C \in C_j \setminus \Lambda_{ij}$ it is that case that $D \to C$ is in $H$. This follows by application of R4 to A, B, C, and D if $A - D$. If $D \to A$ then $D \to B$ by R2 and $D \to C$ by R2. If $A \to D$ then $D \to C$ by R1.□

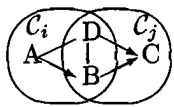

Figure 2: Schematic for Lemma 6

**Lemma 7** Let $T$ be a join tree for a partially oriented chordal graph $H$ without any unshielded colliders and with orientations closed under rules R1, R2, R3, and R4. (i) If $\epsilon_T(C_i, C_j)$ then for all $k$ such that the (unique) path $p$ between $C_i$ and $C_k$ in $T$ is through $j$ then $\epsilon_T(C_i, C_k)$ and (ii) if $C_l$ and $C_m$ are adjacent on the path $p$ then $\gamma(C_l, C_m)$.

**Proof** — Part (i) is proved by induction on length of path between $C_j$ and $C_k$ in the join tree for T. The base case ($j = k$) is trivial and apply Lemma 6 for induction step. Part (ii) follows in a similar fashion. Consider the minimal element $C_l$ of $\epsilon_T$ such that $\epsilon_T(C_l, C_i)$ or $C_l = C_i$. Let $C_m$ be an arbitrary clique such that $\epsilon_T(C_l, C_m)$ and $\Lambda_{lm} \ne \emptyset$. It must be the case that $\gamma(C_l, C_m)$ otherwise it would not be the case that $\epsilon_T(C_l, C_m)$. Then we simply apply Lemma 6 to extend the chain of $\gamma$ between adjacent cliques.□

A partial order $\pi$ over vertices is *compatible* with the orientations in graph $H$ if and only if for no pair of vertices $A$ and $B$ such that $A \to B$ in $H$ is it the case that $\pi(B, A)$.

**Lemma 8** Let $T$ be a join tree for a partially oriented chordal graph $H$ without any unshielded colliders and with orientations closed under rules R1, R2, R3, and R4. (i) there exists a tree ordering which extends $\epsilon_T$, (ii) for all tree orderings $\pi$ which extend $\epsilon_T$ it is the case that $\prec_\pi$ is compatible with H.

**Proof** — (i) Since $\epsilon_T$ is a partial order there is a minimal element. Choose any minimal element as the root of the tree order. By Lemma 7, a tree order constructed in such a manner extends $\epsilon_T$. (ii) Let $\pi$ be a tree order which extends $\epsilon_T$. Suppose that $\prec_\pi$ is *not* compatible with $H$. Then there exists a pair of vertices $A$ and $B$ such that $A \to B$ in $H$ and $B \prec_\pi A$. Let $C_j$ be a clique which contains both $A$ and $B$. For $B \prec_\pi A$ to hold it must be the case that there is a $i$ such that $\pi(C_i, C_j)$. By Lemma 6 $B \to A$. Contradiction.□

**Theorem 4** Let $\mathcal{K}$ be a set of background knowledge consistent with pattern $\Pi$. The result of applying rules R1, R2, R3 and R4 (and orienting edges according to $\mathcal{K}$) to a pattern of some directed acyclic graph is a maximally oriented graph with respect to $\mathcal{K}$.

**Proof** — Let $\Pi_0$ be the result of applying the orientation rules R1, R2, and R3 to the partially directed graph $\Pi$. Given Lemma 1 no orientation of edges not oriented in $\Pi_0$ will create a cycle which includes an edge or edges oriented in $\Pi_0$ and no orientation of an edge not oriented in $\Pi_0$ can create an unshielded collider with an edge oriented in $\Pi_0$. Consider the undirected graph $H$, a subgraph of $\Pi_0$, obtained by removing all of the oriented edges in $\Pi_0$. $H$ is a union of disconnected chordal graph(s); suppose this is not the case. Then, by Lemma 2 all total orderings of the vertices leads to a new unshielded collider, say $\langle A, B, C \rangle$, in $H$. By Lemma 1, the triple $\langle A, B, C \rangle$ also forms an unshielded triple in $\Pi_0$, that is $A \notin adj(C)$ in $\Pi_0$. This is a contradiction; by assumption graph $\Pi$ and thus $\Pi_0$ have all unshielded colliders oriented and that there is an acyclic orientation of the graph $\Pi$ with no new unshielded colliders. Let $\Pi_1$ be the result of orienting all of the edges in $\Pi_0$ that can be oriented with background knowledge and let $\Pi_2$ be the result of applying orientation rule R1, R2, R3, and R4 exhaustively to $\Pi_1$. Let $A - B$ be unoriented in $\Pi_2$ and show that there exists consistent orderings $\alpha$ and $\gamma$ such that $A \to B$ in $H_\alpha$ and $B \to A$ in $H_\gamma$.

*Case 1* — For all $i$ either $A \in C_i \wedge B \in C_i$ or $A \notin C_i \wedge B \notin C_i$. Let T be a join tree for $H$ and let $\pi$ be a tree ordering of T which extends $\epsilon_T$; that one exists follows from Lemma 8. $A$ and $B$ are not comparable with respect to $\prec_\pi$ thus by Lemma 4 we simply choose two extensions (consistent with the ordering existing in $\Pi_2$) of $\prec_\pi$; one with $A \prec_\pi B$ and another with $B \prec_\pi A$. By Lemma 4 we are done.

*Case 2* — There exists an $i$ such that $A \notin C_i \vee B \notin C_i$ and $A \in C_i \vee B \in C_i$. Without loss of generality assume that $A \notin C_i \wedge B \in C_i$. Given that there is an edge between $A$ and $B$ there is a $j$ such that $j \ne i$ and $A \in C_j \wedge B \in C_j$. Since the edge between $A$ and $B$ is unoriented we know that it is not the case that $\gamma(C_i, C_j)$ and thus it is not the case that $\epsilon_T(C_i, C_j)$. Thus the tree order obtained from by letting $C_j$ to be the root of the tree is compatible with $H$ by Lemma 8 Let $\pi$ be the tree ordering obtained by letting $C_j$ to be the root of the tree. Note that for all pairs of vertices in the root clique of the tree ordering are not ordered in the partial order induced by the tree ordering. Let the total order $\prec_1$ be an extension of $\prec_\pi$ consistent with the orientations in $\Pi_2$ such that $A \prec_1 B$ and let the total order $\prec_2$ be an extension of $\prec_\pi$ consistent with the orientations in $\Pi_2$ such that $B \prec_1 A$. Apply Lemma 4 and we are done.□